\def\year{2020}\relax
\documentclass[letterpaper]{article}
\usepackage{sskd}
\usepackage{times}
\usepackage{helvet}
\usepackage{courier}
\usepackage[hyphens]{url}
\usepackage{graphicx}
\usepackage{graphicx}
\usepackage{xcolor}
\usepackage{amsmath,amssymb,amsfonts,dsfont,bm,mathrsfs}
\usepackage{booktabs,multirow,adjustbox}
\usepackage{caption}
\newcommand{\citet}[1]{\citeauthor{#1}~\shortcite{#1}}
\newcommand{\citep}{\cite}

\newcommand{\x}{{\rm\bf x}}      % notation of image
\newcommand{\f}{{\rm\bf f}}      % notation of feature
\newcommand{\loss}{{\mathcal L}} % notation of loss
\nocopyright

\title{An Embarrassingly Simple Approach for Knowledge Distillation}

\author{
  \Large \textbf{Mengya Gao},\textsuperscript{\rm 1$\ast$}
  \Large \textbf{Yujun Shen},\textsuperscript{\rm 2$\ast$}
  \Large \textbf{Quanquan Li},\textsuperscript{\rm 3}
  \Large \textbf{Junjie Yan},\textsuperscript{\rm 3} \\
  \Large \textbf{Liang Wan},\textsuperscript{\rm 1}
  \Large \textbf{Dahua Lin},\textsuperscript{2}
  \Large \textbf{Chen Change Loy},\textsuperscript{4}
  \Large \textbf{Xiaoou Tang}\textsuperscript{2} \\
  \textsuperscript{1}Tianjin University \quad
  \textsuperscript{2}CUHK - SenseTime Joint Lab, The Chinese University of Hong Kong \\
  \textsuperscript{3}SenseTime Group Limited \quad
  \textsuperscript{4}NTU \\
  \{daisy, lwan\}@tju.edu.cn \quad
  \{sy116, dhlin, xtang\}@ie.cuhk.edu.hk \\
  \{liquanquan, yanjunjie\}@sensetime.com \quad
  ccloy@ntu.edu.sg
}

\begin{document}

\maketitle

%%%% Abstract
\begin{abstract}
%%%%
Knowledge Distillation (KD) aims at improving the performance of a low-capacity student model by inheriting knowledge from a high-capacity teacher model.
Previous KD methods typically train a student by minimizing a task-related loss and the KD loss simultaneously, using a pre-defined loss weight to balance these two terms.
In this work, we propose to first transfer the backbone knowledge from a teacher to the student, and then only learn the task-head of the student network.
Such a decomposition of the training process circumvents the need of choosing an appropriate loss weight, which is often difficult in practice, and thus makes it easier to apply to different datasets and tasks.
Importantly, the decomposition permits the core of our method, Stage-by-Stage Knowledge Distillation (SSKD), which facilitates progressive feature mimicking from teacher to student.
Extensive experiments on CIFAR-100 and ImageNet suggest that SSKD significantly narrows down the performance gap between student and teacher, outperforming state-of-the-art approaches.
We also demonstrate the generalization ability of SSKD on other challenging benchmarks, including face recognition on IJB-A dataset as well as object detection on COCO dataset.
\end{abstract}
{\let\thefootnote\relax\footnote{{$\ast$ denotes equal contribution by the authors.}}}

%%%% Section: Introduction
\section{Introduction}\label{sec:introduction}
\begin{figure}[t]
  \centering
  \includegraphics[width=0.85\linewidth]{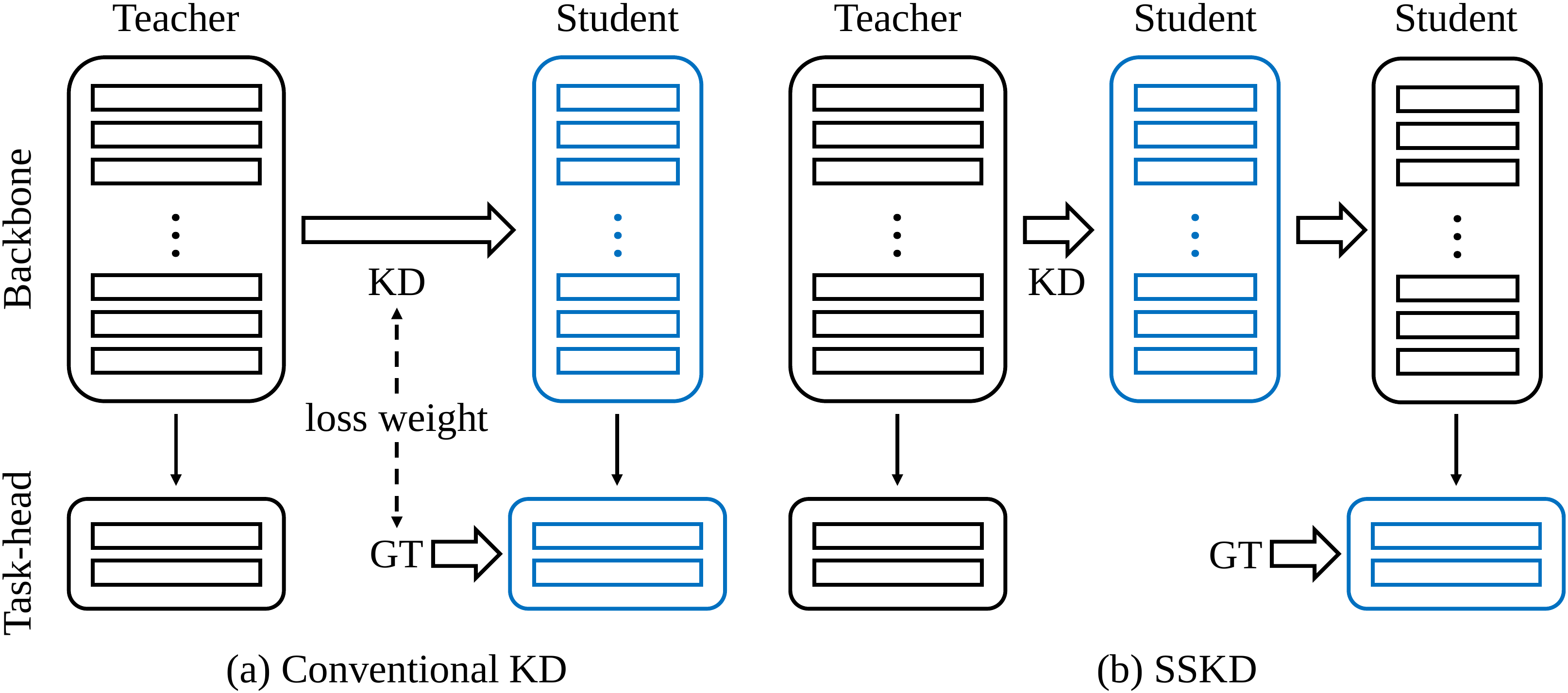}
  \vspace{-5pt}
  \captionsetup{font=small}
  \caption{
    (a) illustrates the conventional KD method, where student learns from teacher and the ground-truth (GT) simultaneously.
    A loss weight is employed to balance these two objects.
    On the contrary, SSKD in (b) proposes to decompose the training of student into two phases, \emph{i.e.}, (i) learning backbone by distilling knowledge from teacher and (ii) learning task-head with ground-truth by fixing backbone.
    Rectangles in \textbf{\textcolor{blue}{blue}} indicate the trainable parts.
  }
  \label{fig:diagram}
  \vspace{-10pt}
\end{figure}

%%%%
Knowledge Distillation (KD) \cite{Hinton2014Distilling}, which allows a small model to achieve competitive performance, is gaining ground in recent years, especially in industrial applications.
The goal of KD is to use a large model, known as \emph{teacher}, to guide the training process of a small one, termed as \emph{student}, such that they can finally produce similar prediction.
Many attempts have been made to improve the above process \cite{Romero2015FitNets,Zagoruyko2017Paying,Yim2017A,lee2018self}, typically treating KD as a regularizer \cite{Hinton2014Distilling,Romero2015FitNets,Huang2017Like}.
In particular, the student is trained to achieve accuracy on the main task and at the same time emulate the behavior of the teacher by minimizing both the main task loss and KD loss, as shown in Fig.\ref{fig:diagram} (a).
A loss weight is usually introduced to balance these two goals.

%%%% Figure: Loss Weight
\begin{figure}[t]
  \centering
  \includegraphics[width=0.9\linewidth]{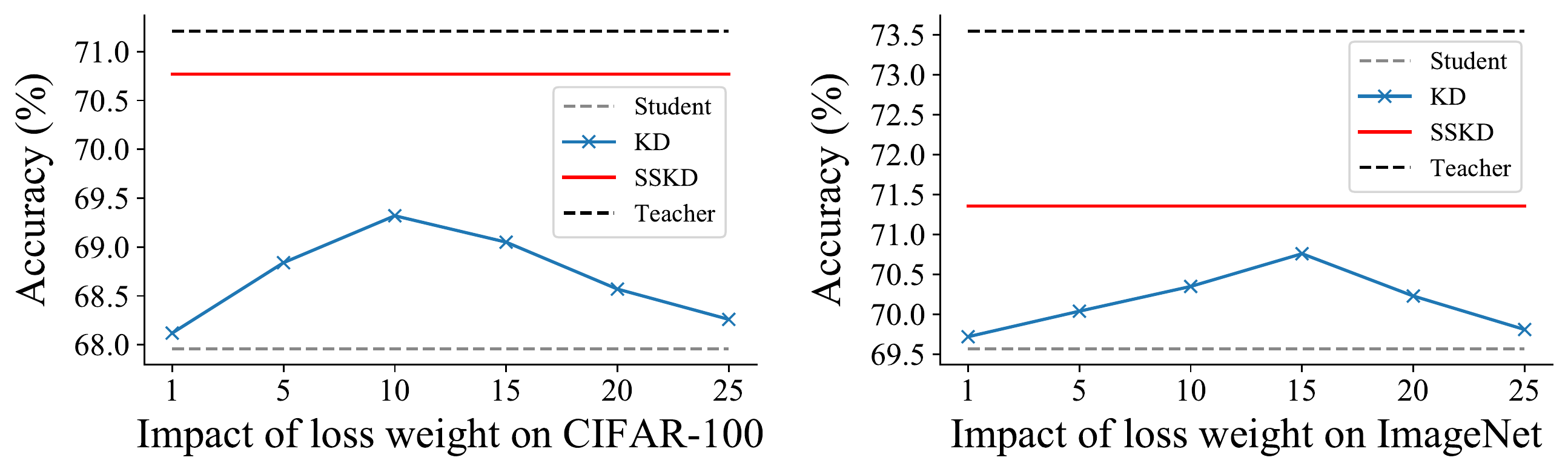}
  \vspace{-8pt}
  \captionsetup{font=small}
  \caption{
    Compared to KD \cite{Hinton2014Distilling}, whose performance is easily affected by loss weight (\textbf{\textcolor{blue}{blue}} line), the proposed SSKD bypasses such hyper-parameter and achieves more competitive and stable results (\textbf{\textcolor{red}{red}} line).
  }
  \label{fig:loss-weight}
  \vspace{-10pt}
\end{figure}

However, choosing an appropriate loss weight is critical but difficult.
On one hand, the performance is sensitive to this hyper-parameter.
We conducted experiments by taking the original KD method \cite{Hinton2014Distilling} as an instance.
As shown in Fig.\ref{fig:loss-weight}, when training ResNet-20 \cite{he2016deep} to learn from ResNet-56 on CIFAR-100 dataset \cite{Krizhevsky2009Learning}, the Top-1 accuracy exhibits around 1.2\% fluctuation with loss weight changing from 1 to 25.
On the other hand, the best option of this value varies over different datasets, tasks, or teacher-student settings.
As shown in Fig.\ref{fig:loss-weight}, the best loss weight on CIFAR-100 is 10, while the best choice for ResNet-18 mimicking ResNet-34 on ImageNet dataset \cite{deng2009imagenet} is 15.
To find the optimal option, it often requires to launch multiple training processes on different choices, which is too costly.

In this work, we propose a unified, simple yet practical solution to reformulate the KD process.
Our approach is motivated by the observation that a model can be inherently divided into two parts, namely (i) a \emph{backbone} for extracting features from the input data, and (ii) a \emph{task-head} for connecting the features to the given task, \emph{e.g.}, classification in a specific domain.
Accordingly, the training of student falls into two phases, as shown in Fig.\ref{fig:diagram} (b).
First, we perform layer-wise feature mimicking to transfer the knowledge from the backbone of teacher to that of student \emph{without} using the ground-truth.
Second, we train the task-head of student with task-dependent loss with \emph{fixed} well-trained backbone.

By decomposing knowledge transfer from task fulfillment, our method circumvents the choice of the loss weight, thus making it easier to apply to various datasets and tasks.
We will show in our experiments that the decomposition of distilling knowledge from teacher and learning from ground-truth generalizes well across a number of different settings.

%%%% Figure: Diagram

To achieve effective distillation on the knowledge contained in the backbone, we further present Stage-by-Stage Knowledge Distillation (SSKD) to transfer knowledge from teacher to student progressively at feature level.
Instead of mimicking features at all levels altogether, we divide the distillation process into multiple stages and only train one sub-network at a time.
We show in the experiments that stage-wise distillation scheme results in more effective transfer of knowledge than single-step distillation.
Although trained with many stages, SSKD can be performed in an efficient way (\emph{i.e.}, not increasing the training time) but still attain good performance.
This benefits from the fact that each stage only trains one part of the network while keeps the remaining parts fixed.

To summarize, our \textbf{contributions} are three-fold:
\begin{itemize}
  \item We propose to decompose the knowledge distillation process into two phases, which alleviates the need of choosing a loss weight yet maintains competitive performance. Taking this advantage, the approach can be stably extended to different datasets and tasks.
  \item We present Stage-by-Stage Knowledge Distillation (SSKD), which is very easy to implement. Even simple, it surpasses state-of-the-art methods on CIFAR-100 and ImageNet benchmarks. For example, on the setting of using ResNet-18 to emulate ResNet-34, SSKD raises the accuracy of student by 1.8\% (compared to 1.2\% obtained by the second competitor) on ImageNet classification task. We further conduct evaluations on other challenging tasks, \emph{e.g.}, large-scale face recognition, and obtain encouraging improvements, demonstrating the generalization ability of SSKD.
  \item We provide new benchmarks by applying knowledge distillation on COCO object detection task \cite{Lin2014Microsoft}. We carry out experiments on two state-of-the-art detection methods, \emph{i.e.}, two-stage FPN method \cite{Lin2017Feature} and one-stage RetinaNet method \cite{lin2017focal}, and achieve significant improvements with both approaches. Specially, with ResNet-152 and ResNet-50 as teacher-student pair, SSKD advances the Average Precision (AP) of original student models from 37.7\% to 40.5\% with FPN, and from 35.9\% to 38.7\% with RetinaNet.
\end{itemize}

%%%% Section: Related Work
\section{Related Work}\label{sec:related-work}
%%%%
The preliminary view of teacher supervising student was adopted by \citet{Ba2014Do}, which trains a shallow network with data labeled by a deep one.
\citet{Hinton2014Distilling} introduced the concept of Knowledge Distillation (KD), which describes the dark knowledge as the soft label produced by teacher model, and also proposed to raise the temperature in softmax function to further distill the knowledge from classification networks.
\citet{Romero2015FitNets} employed the output feature of an intermediate layer in teacher model as hint for student.
\citet{Zagoruyko2017Paying} attempted to transfer spatial attention map, which is defined as the average of feature maps across the channel dimension.
\citet{Huang2017Like} trained student to learn the feature map from teacher through Maximum Mean Discrepancy (MMD), which can be regarded as a sample-based metric to measure the distance between two probability distributions.
\citet{Yim2017A} proposed Flow of Solution Procedure (FSP) to make student mimic the information flow of teacher, which is defined as the Gram matrix of two hidden feature maps.
\citet{lee2018self} improved this idea with Singular Value Decomposition (SVD).

There are two main \textbf{differences} that distinguish SSKD from the above teacher-student training manner.
(i) Existing methods typically trained KD task and the original task at the same time with the help of a loss weight to balance the trade-off.
Even though FitNets \cite{Romero2015FitNets} and FSP \cite{Yim2017A} proposed to offer student with a better initialization by borrowing knowledge from teacher, such information gradually disappears during the following main task learning progress, which is pointed out by \citet{lee2018self}.
Unlike them, SSKD circumvents the requirement of the loss weight hyper-parameter, resulting in \emph{stronger robustness and higher generalization ability}.
(2) SSKD trains student to emulate teacher by mimicking the hidden features \emph{stage by stage}.
In this way, student is capable of learning the knowledge in teacher model more thoroughly, significantly narrowing down the performance gap.
A very recent work, TAKD \cite{mirzadeh2019improved}, also proposed to transfer knowledge from teacher to student gradually by introducing teaching assistant.
Different from their using auxiliary model to facilitate the transferring process, we use different stages to improve the performance.

Besides, KD is also combined with other techniques, such as reinforcement learning \cite{Ashok2018N2N} and adversarial network \cite{Belagiannis2018Adversarial,xu2018training}, to improve the performance.
Some other work blended KD with other model compression methods, \emph{e.g.}, network pruning \cite{Wang2018ProgressiveBK} and weight quantization \cite{han2015deep}, to make student smaller.
\citet{chen2017learning} and \citet{li2017mimicking} are specially designed to distill knowledge from models in objective detection task.
There are also approaches that preform KD without using the teacher-student pair, such as self-learning mechanism \cite{furlanello2018born,lan2018self} and the idea of on-the-fly distillation \cite{zhang2018deep,anil2018large,zhu2018knowledge}.
They focus on different applications of knowledge distillation from ours.

%%%% Figure: Framework
\begin{figure*}[t]
  \centering
  \includegraphics[width=0.9\linewidth]{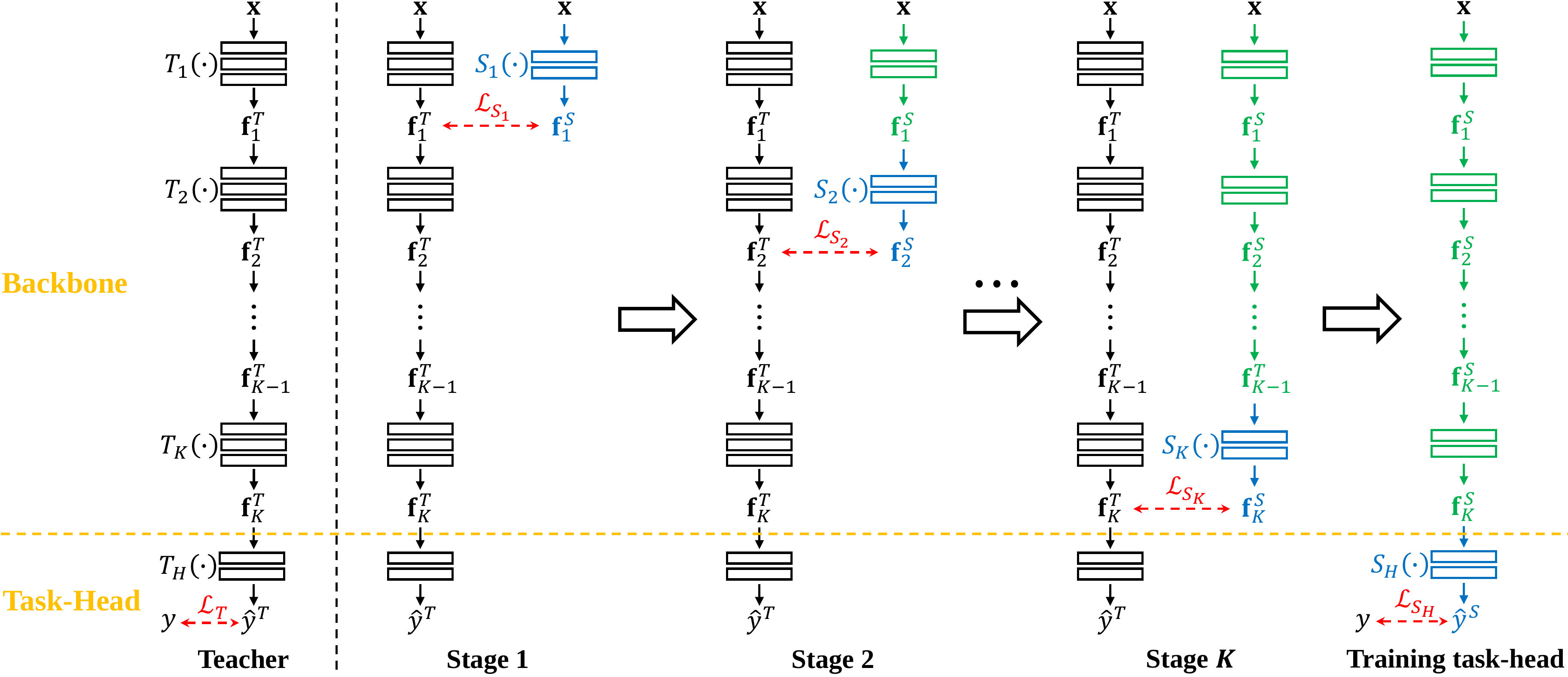}
  \vspace{-5pt}
  \captionsetup{font=small}
  \caption{
    Overview of the proposed SSKD.
    Both teacher and student can be separated into a backbone with $K$ stages and a task-head.
    Before knowledge transfer, teacher model is trained with labeled data (first column).
    After teacher is ready, student first mimics the output feature of teacher progressively, and then learns the task-head with supervision from ground-truth label \emph{without} taking teacher as reference any more (last column).
    Stages in \textbf{\textcolor{blue}{blue}} indicate the current training stage, while all previous stages in \textbf{\textcolor{green}{green}} are fixed.
    Dashed arrows in \textbf{\textcolor{red}{red}} show the loss function for each stage.
  }
  \label{fig:framework}
  \vspace{-10pt}
\end{figure*}

%%%% Section: Stage-by-Stage Knowledge Distillation
\section{Stage-by-Stage Knowledge Distillation}\label{sec:sskd}
%%%%
Fig.\ref{fig:framework} illustrates the training process of our proposed SSKD.
To summarize, a model can be divided into two parts, which are feature extractor (backbone) and data adaptor (task-head).
Based on this separation, the learning of student also follows two steps.
First, the backbone is trained to mimic the output feature of teacher, which is when knowledge distillation actually happens.
Second, the task-head is optimized to fit the ground-truth data, while the parameters in backbone are fixed.
There are no sensitive hyper-parameters such as loss weight in SSKD, making it robust to different datasets and tasks.
Besides training KD task and main task independently, SSKD also improves conventional KD methods through progressive distillation.
More details will be discussed in the following sections.

%%%% Sub-Section: Basics of Knowledge Distillation
\subsection{Basics of Knowledge Distillation}\label{subsec:basics-of-knowledge-distillation}
%%%%
In general, a CNN model $M(\cdot)$ can be treated as two parts, which are the backbone part $M_B(\cdot)$ to extract feature as well as the task-head part, $M_H(\cdot)$ to map the feature to final prediction.
More specifically, given an image-label pair $(\x, y)$, $M(\cdot)$ aims at producing $\hat{y}=M_H(M_B(\x))$, with $\f=M_B(\x)$ as the intermediate feature, to minimize the distance between $y$ and $\hat{y}$
\begin{align}
  \min_{\Theta_M}\loss_M = \phi(y, \hat{y}), \label{eq:model-loss}
\end{align}
where $\Theta_M$, consisting of $\Theta_{M_B}$ and $\Theta_{M_H}$, denotes the trainable parameters of the entire model.
$\phi(\cdot, \cdot)$ is the task-related energy function, such as the softmax cross-entropy loss in classification task and the bounding box regression loss in detection task.

When it comes to KD problem, we have two different models, called teacher $T(\cdot)$ and student $S(\cdot)$ respectively.
Similar as above, $T$ can be formulated as $T_H \circ T_B$ and $S$ as $S_H \circ S_B$, where $\circ$ indicates the function composition.
Correspondingly, the hidden feature and final prediction are denoted as $\{\f^T, \hat{y}^T\}$ and $\{\f^S, \hat{y}^S\}$.
The key challenge is to find out the knowledge contained in teacher model with $\sigma(\f^T, \hat{y}^T)$ and then transfer it to the student.
Prior work typically trains KD task together with the main task in Eq.\eqref{eq:model-loss} with
\begin{align}
  \min_{\Theta_S}\loss_S = \phi(y, \hat{y}^S) + \lambda\psi(\sigma(\f^T, \hat{y}^T), \sigma(\f^S, \hat{y}^S)), \label{eq:mimic-loss}
\end{align}
where $\psi(\cdot, \cdot)$ represents the loss function for knowledge transfer and $\lambda$ is the loss weight to balance these two terms.
Here, the parameters of backbone $\Theta_{S_B}$ and task-head $\Theta_{S_H}$ are updated simultaneously.

%%%% Sub-Section: Training Decomposition
\subsection{Training Decomposition}\label{subsec:training-decomposition}
%%%%
To make the method more flexible to tasks and datasets, we decompose the training of $\Theta_{S_B}$ and $\Theta_{S_H}$.
Specially, we focus the KD task on transferring knowledge contained in the backbone structure, and then only optimize the task-head part with ground-truth label $y$.
This separation can be formulated as
\begin{align}
  \min_{\Theta_{S_B}}\loss_{S_B} &= \psi(\overline{\sigma}(\f^T), \overline{\sigma}(\f^S)), \label{eq:mimic-feature-loss} \\
  \min_{\Theta_{S_H}}\loss_{S_H} &= \phi(y, \hat{y}^S), \label{eq:mimic-label-loss}
\end{align}
where $\overline{\sigma}(\cdot)$ indicates the backbone knowledge only, which differs from $\sigma(\cdot, \cdot)$ in Eq.\eqref{eq:mimic-loss}.
Note that when training either part of $\{\Theta_{S_B}, \Theta_{S_H}\}$, the other part is fixed.
In this work, we choose $\overline{\sigma}(\cdot)$ as identity function, \emph{i.e.}, $\overline{\sigma}(a)=a$, and $\psi(\cdot, \cdot)$ as $l_2$ distance, \emph{i.e.}, $\psi(a, b)=||a-b||_2^2$.

There are two advantages in doing so.
First, despite the discrepancy between the learning capacities of teacher and student in backbone part, they share same structure in task-head, as shown in Fig.\ref{fig:framework}.
Taking image classification task as an example, the final fully-connected layer is viewed as task-head.
Number of parameters in this layer should be feature dimension times number of categories.
Thus, as long as student share same shape of feature map with teacher, their task-heads should have similar modeling ability.
From this point of view, there is no need to transfer knowledge in this part.
Second, by decomposing the training of backbone and task-head, we do not depend on the loss weight ($\lambda$ in Eq.\eqref{eq:mimic-loss}), making our method more extendable and stable.

%%%% Sub-Section: Stage-by-Stage Training Scheme
\subsection{Stage-by-Stage Training Scheme}\label{subsec:stage-by-stage-training-scheme}
%%%%
From the discussion above, it is crucial for student to capture similar features encapsulated by the teacher.
To this end, we break down both teacher and student into multiple stages and let the student to mimic features of the teacher progressively in a stage-wise manner, as shown in Fig.\ref{fig:framework}.
Taking the teacher model as an example, we have
\begin{align}
  \left\{\begin{aligned}
    \f_0^T &= \x, \\
    \f_i^T &= T_i(\f_{i-1}^T), \quad i = 1,2,\cdots,K, \\
    T_B &= T_{K} \circ T_{K-1} \circ \cdots \circ T_1,
  \end{aligned}\right. \label{eq:stage}
\end{align}
where $K$ is the total number of stages contained in the backbone structure.
$\f_i^T$ is the output feature of the $i$-th stage, while $\f_0^T$ is the initial feature (\emph{i.e.}, the input image).
Similarly, $S_B(\cdot)$ is also divided into $K$ stages.
Under such separation, the feature transfer process as shown in Eq.\eqref{eq:mimic-feature-loss} can be further split apart as follows
\begin{align}
  \min_{\Theta_{S_i}}\loss_{S_i} = \psi(\overline{\sigma}(\f_i^T), \overline{\sigma}(\f_i^S)), \quad i = 1,2,\cdots,K. \label{eq:stage-loss}
\end{align}

Specifically, when training a new stage of student, we fix all parameters in previous stages to prevent the transferred knowledge from vanishing \cite{lee2018self}.
Another advantage in doing so is to speed up the training process, since the gradient back-propagation is only applied to a subset of parameters in each stage.
Furthermore, although the stages are trained separately, they are not completely independent.
That is because each stage will take the feature produced by the previous stage as input.
%
% In other words, suppose the $i$-th stage of student cannot exactly reproduce the same feature as teacher, such difference between $\f_i^S$ and $\f_i^T$ can be further handled by the $(i+1)$-th stage of student.
%
In this way, each stage gets substantial learning from teacher, but also complements with each other to achieve better performance.

%%%% Sub-Section: Implementation
\subsection{Implementation}\label{subsec:implementation}
%%%%
We first introduce the way to perform the stage partition.
In general, we treat each resolution down-sampling layer as a breakpoint.
Taking ResNet family \cite{he2016deep} as an example, the input image is with size $224\times224$.
Excluding the first convolution layer and pooling layer, there are mainly four sets of residual blocks, each of which produces an intermediate feature map.
The spatial resolutions are $56\times56$, $28\times28$, $14\times14$, $7\times7$, respectively.
In our implementation, we use these four feature maps to compute the stage-wise mimicking loss in Eq.\eqref{eq:stage-loss}.

Since we are using $l_2$ distance, stage features from student and teacher are required to have the same dimension.
This assumption has been widely practiced in previous studies \cite{Romero2015FitNets,Zagoruyko2017Paying,Yim2017A}.
We also use some tricks to deal with the shape mismatch cases.
Specifically, if the feature maps from teacher and student have different number of channels, we just add an additional convolution layer with $1\times1$ kernel size to student such that the $l_2$ distance can be applied.
Here, the added $1\times1$ kernel is only used in the training phase for loss computation.
If they have different spatial resolutions, we just resize the student's feature map, as described in \citet{Zagoruyko2017Paying}.

When training a particular stage, we use SGD optimizer with momentum equal to 0.9.
The learning rate is set to 0.01 initially and decreased 10\% every time the feature distance $||\psi(\f_i^T, \f_i^S)||_2$ does not decrease any more.
When the learning rate achieves $1e^{-5}$, we suppose the current stage is well trained and start training the next stage.
Taking experiments on ImageNet as an example, we train 60 epochs for each stage and decrease the learning rate with multiplier of 0.1 at 18-th, 36-th, 54-th epoch.
Recall that when training a particular stage in SSKD, other stages are fixed.
After the backbone is sufficiently optimized, the task-head is trained with Eq.\eqref{eq:mimic-label-loss}, which is task-dependent.
For example, the cross-entropy loss (\emph{i.e.}, $\phi(y, \hat{y}) = -\sum_{i=1}^N y_i\log\hat{y}_i$ where $N$ is the number of categories) activated by softmax function is used for classification task.

%%%% Table: Separating Training
\setlength{\tabcolsep}{6pt}
\begin{table}[b]
  \vspace{-8pt}
  \captionsetup{font=small}
  \caption{Ablation study on separate training of backbone and task-head on ImageNet dataset.}
  \label{tab:separate-training}
  \vspace{-7pt}
  \centering\small
  \begin{tabular}{lccc}
    \toprule
    \textbf{Method} & \textbf{Model} & \textbf{Top-1} & \textbf{Top-5} \\
    % \midrule
    % student & ResNet-18 & 69.572 & 89.244 \\
    % student (fixed backbone) & ResNet-18 & 69.952 & 89.242 \\
    \midrule
    Teacher (end-to-end)     & ResNet-34 & 73.55 & 91.46 \\
    Teacher (fixed backbone) & ResNet-34 & 73.53 & 91.44 \\
    \midrule
    KD (end-to-end)          & ResNet-18 & 70.76 & 89.81 \\
    KD (fixed backbone)      & ResNet-18 & 70.75 & 89.84 \\
    \bottomrule
  \end{tabular}
\end{table}

%%%% Section: Experiments
\section{Experiments}\label{sec:experiments}
%%%%
In this section, we first analyze SSKD with ablation studies.
Then, to further evaluate the performance of SSKD, we carry out extensive experiments on various challenging tasks.
In particular, we demonstrate the superiority of SSKD over the state-of-the-art methods on CIFAR-100 \cite{Krizhevsky2009Learning} and ImageNet \cite{deng2009imagenet} datasets.
We also demonstrate the strong generalization ability of SSKD by applying it on IJB-A \cite{klare2015pushing} and CASIA-WebFace \cite{yi2014learning} face recognition benchmarks as well as COCO \cite{Lin2014Microsoft} detection benchmark.

%%%% Sub-Section: Ablation Study
\subsection{Ablation Studies}\label{subsec:ablation-study}
%%%%

\noindent\textbf{Separate Training of Backbone and Task-head.}
SSKD proposes to train backbone and task-head separately to circumvent the choice of loss weight.
In this part, we set up a series of experiments on ImageNet dataset to validate that training the task-head \emph{after} (instead of concurrent with) the backbone will not affect the performance.
On ImageNet dataset, we first train the backbone and task-head of ResNet-34 simultaneously with task loss, then randomly re-initialize the weights of the final fully-connected layer, and finally fix the backbone and only train the fully-connected layer with cross-entropy loss.
The performances of end-to-end training and separate training are shown in Tab.\ref{tab:separate-training}, where they achieve almost same results.
We can conclude that, as long as the model is capable of extracting discriminative feature from the input image, it doesn't matter whether the task-head is trained separately from or together with the backbone.

To further show that this conclusion is also applicable to knowledge distillation task, we conduct experiments on KD method \cite{Hinton2014Distilling}.
Here we use ResNet-18 as the student model and the well-trained ResNet-34 as teacher.
Same as above, we first train student to learn from teacher with both KD loss and task loss, then randomize the fully-connected layer, and finally train the fully-connected layer with task loss only.
As shown in Tab.\ref{tab:separate-training}, KD can also work well when training backbone and task-head separately.
This demonstrates the feasibility of SSKD.

%%%% Table: Stage-by-Stage
\setlength{\tabcolsep}{8pt}
\begin{table}[t]
  \captionsetup{font=small}
  \caption{Ablation experiments of stage-by-stage learning strategy on CIFAR-100 dataset.}
  \label{tab:stage-by-stage}
  \vspace{-8pt}
  \centering\small
  \begin{tabular}{lccc}
    \toprule
    \textbf{\# Stages} & \textbf{Multi-loss Training} & \textbf{SSKD} & \textbf{Improv.} \\
    \midrule
    Student  & 67.96 & -     & -    \\
    Teacher  & 71.21 & -     & -    \\
    \midrule
    1 stage  & 69.24 & 69.24 & 0.00 \\
    2 stages & 69.29 & 70.03 & 0.74 \\
    3 stages & 69.37 & 70.46 & 1.09 \\
    4 stages & 69.46 & 70.77 & 1.31 \\
    \bottomrule
  \end{tabular}
  \vspace{-2pt}
\end{table}

%%%% Figure: Feature
\begin{figure}[t]
  \centering
  \includegraphics[width=0.85\linewidth]{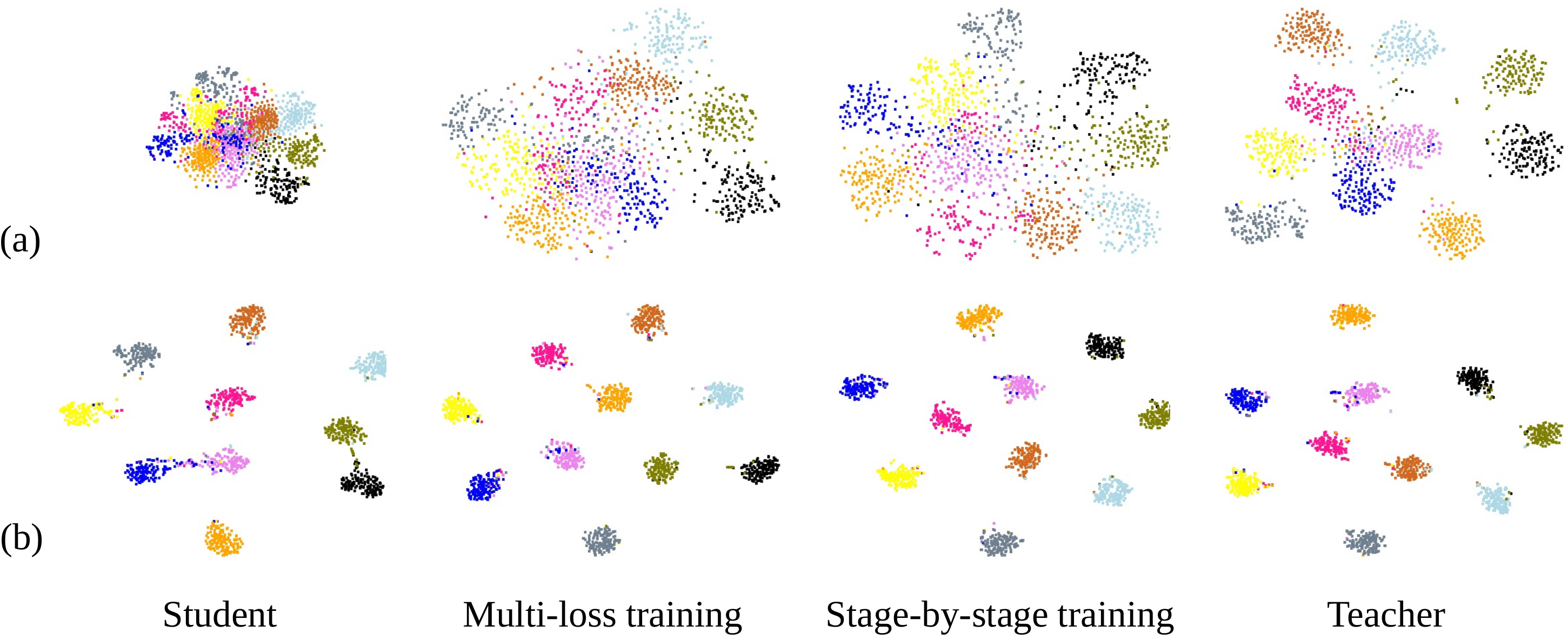}
  \vspace{-5pt}
  \captionsetup{font=small}
  \caption{
    Visualization of features produced by different training schemes, including (a) feature of the 3rd stage and (b) feature of the 4th stage (final stage).
    Best viewed in color.
  }
  \label{fig:feature}
  \vspace{-5pt}
\end{figure}

\noindent\textbf{Comparison with multi-loss training.}
An alternative way for our stage-by-stage training is to sum up the distillation losses from each stage and minimize them together, resulting in multi-loss training.
Tab.\ref{tab:stage-by-stage} shows the comparison results on CIFAR-100 dataset with ResNet-56 and ResNet-20 as the teacher-student pair.
In the experiments, `1 stage' means that only one mimic loss is added at the final feature, where SSKD is identical to multi-loss training.
`2 stages' means separating the backbone network into two parts and and an additional mimic loss from the intermediate layer is introduced, and so on and so forth.
It is evident from Tab.\ref{tab:stage-by-stage} that introducing losses from intermediate features can actually help student get better performance.
Meanwhile, SSKD always outperforms multi-loss training, and the more stages involved, the higher the improvement is.

We also use t-SNE \cite{maaten2008visualizing} as a visualization tool to show that SSKD helps student learn a better representation.
Fig.\ref{fig:feature} visualizes the hidden features of the 3rd stage and 4th stage.
Here, we randomly choose 10 classes out of the 100 classes in CIFAR-100.
We can see that the features from SSKD are more discriminative than those from the multi-loss training manner.

%%%% Table: Number of stages
\setlength{\tabcolsep}{10pt}
\begin{table}[t]
  \captionsetup{font=small}
  \caption{Ablation study on number of stages with CIFAR-100 dataset. Last column shows the Top-1 accuracy improvement compared to training student from scratch (first row).}
  \label{tab:number-of-stages}
  \vspace{-8pt}
  \centering\small
  \begin{tabular}{lccc}
    \toprule
    \textbf{\# Stages} & \textbf{Top-1} & \textbf{Top-5} & \textbf{Top-1 Improv.}\\
    \midrule
    Student  & 67.96 & 90.36 & -    \\
    Teacher  & 71.21 & 91.09 & -    \\
    \midrule
    1 stage  & 69.24 & 90.50 & 1.28 \\
    2 stages & 70.03 & 91.38 & 2.07 \\
    3 stages & 70.46 & 91.47 & 2.50 \\
    4 stages & \textbf{70.77} & \textbf{91.49} & \textbf{2.81} \\
    \midrule
    5 stages & 70.79 & 91.49 & 2.83 \\
    6 stages & 70.82 & 91.61 & 2.86 \\
    7 stages & 70.85 & 91.95 & 2.89 \\
    8 stages & \textbf{70.93} & \textbf{91.96} & \textbf{2.97} \\
    \bottomrule
  \end{tabular}
  \vspace{-8pt}
\end{table}

%%%% Table: Different Teachers
\setlength{\tabcolsep}{5pt}
\begin{table}[t]
  \captionsetup{font=small}
  \caption{Experiments by using student (ResNet-20) to learn from different teachers on CIFAR-100 dataset.}
  \label{tab:different-teachers}
  \vspace{-8pt}
  \centering\small
  \begin{tabular}{lccc}
    \toprule
    \textbf{Teacher} & \textbf{Teacher Top-1} & \textbf{Student Top-1} & \textbf{Improv.} \\
    \midrule
    - & - & 67.96 & - \\
    \midrule
    VGG-16       & 71.03 & 69.98 & 2.02 \\
    ResNet-56    & 71.21 & 70.77 & 2.81 \\
    ResNet-110   & 71.86 & 70.84 & 2.88 \\
    DenseNet-121 & 75.45 & 73.27 & 5.21 \\
    \bottomrule
  \end{tabular}
  \vspace{-10pt}
\end{table}

\noindent\textbf{Number of Stages.}
From Tab.\ref{tab:stage-by-stage}, we observe that more stages can lead to a better performance in the student.
In this part, we further add more stages to see the limitation of SSKD.
Here, based on the original 4-stages partition, `5 stages' means that the first stage is evenly divided into two halves, `6 stages' means that the first two stages are evenly divided, and so on and so forth.
The results are summarized in Tab.\ref{tab:number-of-stages}.
While adding more stages still attain strong performance against the upper-bound teacher model, we note that adding more stages only bring marginal improvements.
For simplicity, we recommend using the resolution down-sampling layers as the stage breakpoints (\emph{i.e.}, 4 stages), which will be used in the following experiments.

\noindent\textbf{Teachers with Different Architectures.}
In practice, student and teacher may not always share the same type of structure.
This requires a KD method to be robust to heterogeneous CNN types.
As mentioned above, SSKD only needs the hidden feature maps from teacher and student to have the same shape, which can be easily extended to different settings.
To verify the generalization ability of SSKD, we employ ResNet-20 to learn from teachers with various widths, depths, and structures, including VGG-16 \cite{simonyan2015very}, ResNet-56, ResNet-110, and DenseNet-121 \cite{huang2017densely}.
Tab.\ref{tab:different-teachers} shows the results, and the improvements over baseline model (ResNet-20 trained from scratch) are reported in the last column.
We obtain two observations:
(i) by using SSKD, the performance of student significantly gets improved in comparison to the baseline no matter whether it shares the similar network structure with the teacher or not;
(ii) the performance of the student consistently improves along with more powerful teacher.
The results suggest the strong robustness of SSKD in handling different architectures.

%%%% Figure: Loss Weight Choice
\begin{figure}[t]
  \centering
  \includegraphics[width=0.95\linewidth]{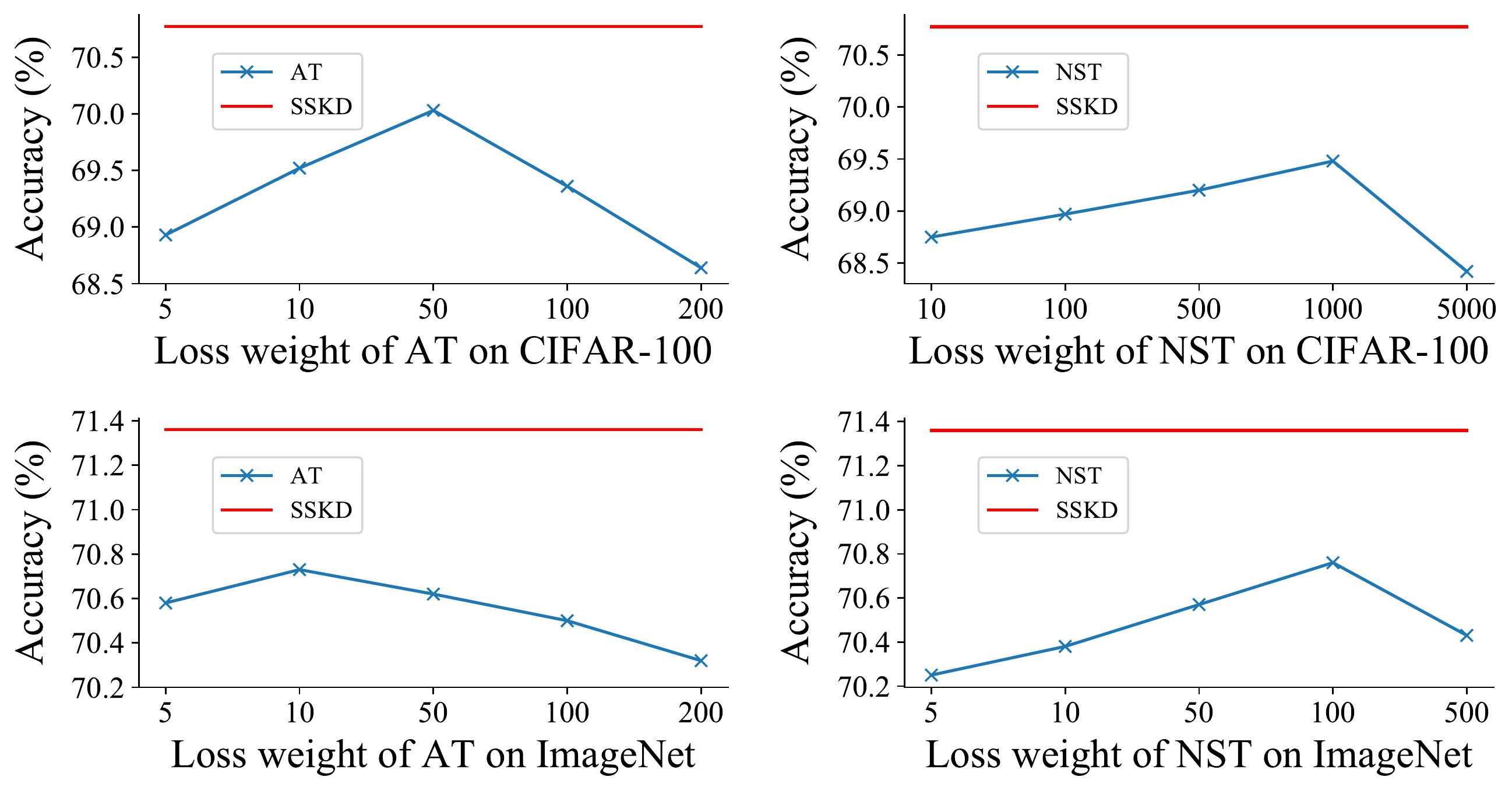}
  \vspace{-5pt}
  \captionsetup{font=small}
  \caption{
    Choices of loss weights.
    For each method on a particular dataset, we conduct a series of experiments by varying the loss weight and then choose the best option.
    Compared to previous approaches (\textbf{\textcolor{blue}{blue}} line), SSKD (\textbf{\textcolor{red}{red}} line) does not require tuning the hyper-parameter yet achieves higher performance.
  }
  \label{fig:loss-weight-choice}
  \vspace{-5pt}
\end{figure}

%%%% Sub-Section: Experiments on Image Classification
\subsection{Experiments on Image Classification}\label{subsec:image-classification}
%%%%
In this section, we evaluate our method on CIFAR-100 and ImageNet datasets.
ResNet-20 and ResNet-56 are employed as student and teacher respectively for CIFAR-100 dataset, while ResNet-18 and ResNet-34 are used for ImageNet dataset.
To further validate the effectiveness of our method, we compare against existing knowledge distillation approaches, including KD \cite{Hinton2014Distilling}, FitNets \cite{Romero2015FitNets}, AT \cite{Zagoruyko2017Paying}, and NST \cite{Huang2017Like}.

\noindent\textbf{Settings for Competitors.}
We set the temperature for softmax function as $T=4$ in KD method.
For FitNets, we use the features with resolution $28\times28$ as the intermediate hint layer.
For AT and NST, we compute the transfer loss with four output features from down-sampling layers, which is the same as SSKD.
Recall that all these methods require loss weight to balance the original task and KD task, namely $\alpha$ in KD\footnote[1]{$\alpha=0.2$ (0.2 \emph{v.s.} 0.8) in KD is equivalent to $\lambda=4$ (1 \emph{v.s.} 4)}, $\lambda$ in FitNets and NST, and the divisor of $\beta$ in AT.
As there is no standard rule on how to choose these hyperparameters for every experimental setting (\emph{i.e.}, task, dataset, and teacher-student pair), we exhaustively find out the best option for each setting by conducting multiple experiments with different loss weights.
Fig.\ref{fig:loss-weight-choice} shows some selection processes.
We can see that existing methods are very sensitive to the value of loss weight.

\noindent\textbf{Training Efficiency.}
Recall that, even SSKD employs four stages, only partial parameters are updated in each stage such that the training of SSKD is efficient.
Taking ResNet-18 mimicking ResNet-34 on ImageNet as an example, we train KD, FitNets, AT, and NST with 100 epochs, and train SSKD with 60 epochs for each stage\footnote[2]{Even other stages are fixed when training a particular stage, the forward propagation still takes some time. Thus, to make fair comparison, we train SSKD with less epochs than other methods such that they share same training time.}.
In this way, using 8 GPUs, we can finish the training of all methods with around 60 hours, resulting in same training time.
In other words, SSKD consumes same computational resources as other approaches at both training and inference phases, leading to fair comparison.

\noindent\textbf{Evaluation on CIFAR-100.}
CIFAR-100 dataset is widely used in KD task \cite{Romero2015FitNets,Huang2017Like,Yim2017A,lee2018self}.
Tab.\ref{tab:cifar100} shows the comparative results.
Our method surpasses other work in both Top-1 accuracy and Top-5 accuracy.
SSKD even achieves higher Top-5 performance than the teacher model.

\noindent\textbf{Evaluation on ImageNet.}
We also conduct larger-scale experiments on ImageNet dataset \cite{deng2009imagenet}, which includes over 1M training images and 50K testing images collected from 1,000 categories.
As shown in Tab.\ref{tab:imagenet}, our method improves the baseline model with 1.8\% Top-1 accuracy and beats the second competitor by 0.6\% Top-1 accuracy.

\noindent\textbf{Discussion.}
By cross-comparison of Tab.\ref{tab:cifar100} and Tab.\ref{tab:imagenet}, our method show strong robustness across different datasets with drastically different number of categories.
The stability is advantageous to other competitors.
For example, FitNets performs well on CIFAR-100, but is not as good as KD on ImageNet.
Also, NST succeeds on ImageNet, but shows poor performance on CIFAR-100.
As we can see from Tab.\ref{tab:cifar100} and Tab.\ref{tab:imagenet}, KD, FitNets, AT, and NST require different loss weights for different datasets to balance the task loss and distillation loss.
On the contrary, this issue is alleviated in SSKD, making our method more easily to apply to various datasets.
This can also be concluded in Fig.\ref{fig:loss-weight-choice}, where our approach is more stable and achieves better accuracy on both CIFAR-100 and ImageNet datasets.

%%%% Table: CIFAR-100
\setlength{\tabcolsep}{5pt}
\begin{table}[t]
  \captionsetup{font=small}
  \caption{Comparison results of image classification task on CIFAR-100 dataset.}
  \label{tab:cifar100}
  \vspace{-5pt}
  \centering\small
  \begin{tabular}{lcccc}
    \toprule
    \textbf{Method} & \textbf{Structure} & \textbf{Top-1} & \textbf{Top-5} & \textbf{Loss Weight} \\
    \midrule
    Student & ResNet-18 & 67.96 & 90.36 & -    \\
    Teacher & ResNet-34 & 71.21 & 91.09 & -    \\
    \midrule
    KD      & ResNet-18 & 69.32 & 90.13 & 10   \\
    FitNets & ResNet-18 & 69.96 & 90.57 & 100  \\
    AT      & ResNet-18 & 70.03 & 90.84 & 50   \\
    NST     & ResNet-18 & 69.48 & 90.27 & 1000 \\
    \midrule
    Ours    & ResNet-18 & \textbf{70.77} & \textbf{91.49} & - \\
    \bottomrule
  \end{tabular}
  \vspace{-8pt}
\end{table}

%%%% Table: ImageNet
\setlength{\tabcolsep}{5pt}
\begin{table}[t]
  \captionsetup{font=small}
  \caption{Comparative results of image classification task on ImageNet dataset.}
  \label{tab:imagenet}
  \vspace{-5pt}
  \centering\small
  \begin{tabular}{lcccc}
    \toprule
    \textbf{Method} & \textbf{Structure} & \textbf{Top-1} & \textbf{Top-5} & \textbf{Loss Weight} \\
    \midrule
    Student & ResNet-18 & 69.57 & 89.24 & - \\
    Teacher & ResNet-34 & 73.55 & 91.46 & - \\
    \midrule
    KD      & ResNet-18 & 70.76 & 89.81 & 15 \\
    FitNets & ResNet-18 & 70.66 & 89.23 & 10 \\
    AT      & ResNet-18 & 70.73 & 90.04 & 10 \\
    NST     & ResNet-18 & 70.76 & 89.59 & 100 \\
    \midrule
    Ours    & ResNet-18 & \textbf{71.36} & \textbf{90.50} & - \\
    \bottomrule
  \end{tabular}
  \vspace{-5pt}
\end{table}

%%%% Sub-Section: Experiments on Face Recognition
\subsection{Experiments on Face Recognition}\label{subsec:face-recognition}
%%%%
We further conduct experiments on face recognition task, which is challenging due to the tremendous number of categories.
We use CASIA-WebFace \cite{yi2014learning} as training set, which has 494,414 images of 10,575 subjects.
Then, the well-trained model is evaluated on IJB-A dataset \cite{klare2015pushing}, which consists of 25,808 images of 500 subjects.
The open-set setting in face recognition differs drastically from CIFAR-100 and ImageNet, since the training set and testing set have different categories.
We use this setting to validate how SSKD can be generalized from one dataset to another.
According to IJB-A protocol, only the output feature will be used for testing.
Therefore, we do not tune the task-head in this experiment.
We also applied KD \cite{Hinton2014Distilling} on this task for comparison.

Tab.\ref{tab:face-recognition} shows the results.
It is observed that SSKD is capable of improving the performance of student under both verification protocol and identification protocol, suggesting that SSKD indeed helps the student in learning a better representation.
By comparing with KD, which barely increases the performance of student, the proposed SSKD shows much stronger generalization ability to different datasets and tasks.

%%%% Table: Face Recognition
\setlength{\tabcolsep}{1.7pt}
\begin{table}[t]
  \captionsetup{font=small}
  \caption{Experimental results on face recognition task. ResNet-18 and ResNet-50 are employed as student and teacher respectively. All models are trained on CASIA-WebFace dataset, and then evaluated on IJB-A dataset.}
  \label{tab:face-recognition}
  \vspace{-5pt}
  \centering\small
  \begin{tabular}{lcccc}
    \toprule
    & \multicolumn{2}{c}{\textbf{Verification}} & \multicolumn{2}{c}{\textbf{Identification}} \\ \cmidrule(lr){2-3} \cmidrule(lr){4-5}
    \textbf{Methods} & \textbf{@FAR=0.01} & \textbf{@FAR=0.001} & \textbf{@Rank-1} & \textbf{@Rank-5} \\
    \midrule
    Student & $76.9 \pm 4.3$ & $55.2 \pm 5.7$ & $84.6 \pm 2.1$ & $92.4 \pm 1.5$ \\
    Teacher & $83.7 \pm 2.8$ & $59.3 \pm 3.6$ & $90.1 \pm 1.2$ & $95.8 \pm 0.7$ \\
    \midrule
    KD      & $77.2 \pm 4.3$ & $55.4 \pm 5.9$ & $85.3 \pm 1.9$ & $93.0 \pm 1.3$ \\
    SSKD    & $\bm{80.5 \pm 3.0}$ & $\bm{56.9 \pm 4.3}$ & $\bm{86.8 \pm 1.3}$ & $\bm{93.5 \pm 1.1}$ \\
    \bottomrule
  \end{tabular}
  \vspace{0pt}
\end{table}

%%%% Sub-Section: Experiment on Object Detection
\subsection{Experiments on Object Detection}\label{subsec:object-detection}
%%%%
To further validate the generalization ability of SSKD on different tasks, we conducted more experiments on the challenging COCO object detection benchmark \cite{Lin2014Microsoft}, which has 80 object categories.
We use the union of $80K$ train images and a $35K$ subset of val images as training set and evaluate on $5K$ subset of val images (minival) following \cite{Lin2017Feature,lin2017focal}.
Standard COCO AP@0.5-0.95 metric as well as the Average Precision (AP) on different object sizes are used for evaluation.
We experiment by adopting two state-of-the-art methods, FPN \cite{Lin2017Feature} and RetinaNet \cite{lin2017focal}, as two-stage and one-stage detection method respectively.
Implementations exactly follow the original paper \cite{Lin2017Feature,lin2017focal}, and 2$\times$ training schedule setting is used, following Detectron \cite{Detectron2018}, to avoid the potential influence of training steps on the final performance.
Feature pyramid network is treated as the backbone for both FPN and RetinaNet.
For FPN, we take the region proposal network and the fast r-cnn layers (\emph{i.e.}, ROI pooling and fully-connected layers) as task-head.
For RetinaNet, we take the classification and regression subnets as task-head.
We use ResNet-18 and ResNet-50 as students to learn from ResNet-101 and ResNet-152, respectively.
Results are shown in Tab.\ref{tab:detection-fpn} and Tab.\ref{tab:detection-retinanet}.
Here, since KD \cite{Hinton2014Distilling} is specially designed for classification task, it can not be applied to object detection.
Hence, we do not compare with KD in this experiment.

\vspace{2pt}\noindent\textbf{Evaluation on FPN.}
As shown in Tab.\ref{tab:detection-fpn}, SSKD achieved significant improvements compared to original FPN models \cite{Lin2017Feature}.
When using ResNet-101 as the teacher model, SSKD improves ResNet-18 by 2.8\% and ResNet-50 by 2.1\%.
We surprisingly found that, after learning from teacher (ResNet-101) with SSKD, the student (ResNet-50) even achieves higher performance (39.8\% v.s. 39.5\%).
Experiments with a larger teacher model, ResNet-152, also shows consistent improvements.
For example, SSKD improves ResNet-50 with 2.8\% higher AP.

\vspace{2pt}\noindent\textbf{Evaluation on RetinaNet.}
Tab.\ref{tab:detection-retinanet} presents the results with RetinaNet method, where SSKD also obtains large improvements.
By distilling the knowledge in ResNet-101 with SSKD, ResNet-18 and ResNet-50 achieve 3.1\% and 1.3\% increase on AP respectively.
Besides the averaged AP (second column), students learned with SSKD show consistent improvements on average precision for all object sizes.
Meanwhile, when using ResNet-152 as teacher, the performances of ResNet-18 and ResNet-50 are also boosted by 3.2\% and 2.8\% respectively.
Significant improvements of SSKD under different students and teachers settings demonstrate the effectiveness and great generalization ability on object detection task.

%%%% Table: Detection FPN
\setlength{\tabcolsep}{4pt}
\begin{table}[t]
  \captionsetup{font=small}
  \caption{Experimental results on COCO object detection task with FPN method. ResNet-18 and ResNet-50 are employed as students, while ResNet-101 and ResNet-152 are used as teachers.}
  \label{tab:detection-fpn}
  \vspace{-8pt}
  \centering\small
  \begin{tabular}{lcccccc}
    \toprule
    \textbf{Methods} & \textbf{AP} & \textbf{AP$_{50}$} & \textbf{AP$_{75}$} & \textbf{AP$_S$} & \textbf{AP$_M$} & \textbf{AP$_L$} \\
    \midrule
    ResNet-18 (R18) & 33.6 & 55.1 & 35.6 & 18.9 & 36.0 & 44.1 \\
    ResNet-50 (R50) & 37.7 & 57.6 & 40.5 & 19.8 & 41.1 & 49.7 \\
    ResNet-101 (R101) & 39.5 & 61.1 & 42.9 & 23.2 & 43.7 & 50.6 \\
    ResNet-152 (R152) & 41.6 & 63.1 & 45.3 & 25.5 & 46.0 & 53.9 \\
    \midrule
    R18 \emph{mimic} R101 & 36.4 & 57.6 & 39.2 & 19.3 & 40.1 & 49.2 \\
    R18 \emph{mimic} R152 & 36.7 & 57.8 & 39.7 & 19.5 & 40.2 & 50.1 \\
    R50 \emph{mimic} R101 & 39.8 & 61.0 & 43.2 & 23.3 & 44.2 & 50.7 \\
    R50 \emph{mimic} R152 & 40.5 & 61.8 & 44.0 & 23.3 & 44.9 & 52.8 \\
    \bottomrule
  \end{tabular}
  \vspace{-8pt}
\end{table}

%%%% Table: Detection RetinaNet
\setlength{\tabcolsep}{4pt}
\begin{table}[t]
  \captionsetup{font=small}
  \caption{Experimental results on COCO object detection task with RetinaNet method. ResNet-18 and ResNet-50 are employed as students, while ResNet-101 and ResNet-152 are used as teachers.}
  \label{tab:detection-retinanet}
  \vspace{-8pt}
  \centering\small
  \begin{tabular}{lcccccc}
    \toprule
    \textbf{Methods} & \textbf{AP} & \textbf{AP$_{50}$} & \textbf{AP$_{75}$} & \textbf{AP$_S$} & \textbf{AP$_M$} & \textbf{AP$_L$} \\
    \midrule
    ResNet-18 (R18) & 32.4 & 50.8 & 34.1 & 16.6 & 34.8 & 44.6 \\
    ResNet-50 (R50) & 35.9 & 55.3 & 38.2 & 18.4 & 39.5 & 47.8 \\
    ResNet-101 (R101) & 37.6 & 57.1 & 40.2 & 19.8 & 41.3 & 50.5 \\
    ResNet-152 (R152) & 39.9 & 59.7 & 42.8 & 22.2 & 43.9 & 52.9 \\
    \midrule
    R18 \emph{mimic} R101 & 35.5 & 53.7 & 37.4 & 17.3 & 38.2 & 49.4 \\
    R18 \emph{mimic} R152 & 35.6 & 54.2 & 37.5 & 17.9 & 38.8 & 50.1 \\
    R50 \emph{mimic} R101 & 37.2 & 56.9 & 40.0 & 19.5 & 41.2 & 50.2 \\
    R50 \emph{mimic} R152 & 38.7 & 58.0 & 41.6 & 20.3 & 41.8 & 51.6 \\
    \bottomrule
  \end{tabular}
  \vspace{-8pt}
\end{table}

%%%% Section: Conclusion
\section{Conclusion}\label{sec:conclusion}
%%%%
In this work, we argue that first distilling the backbone knowledge from teacher and then fitting the task-head with labeled data can improve the robustness and generalization ability of KD method.
Based on this, we present SSKD by training student to mimic the intermediate features of teacher gradually.
Extensive experiments suggest the significant improvements achieved by SSKD on various tasks.

{\small
  \bibliographystyle{sskd}
  \bibliography{ref}
}

\end{document}